# UQuAD1.0: Development of an Urdu Question Answering Training Data for Machine Reading Comprehension

**Samreen Kazi[1*], Shakeel Khoja[2*]**
[1, 2] School of Mathematics & Computer Science
Institute of Business Administration, Karachi, Pakistan
{sakazi, skhoja}@iba.edu.pk

**Abstract**
In recent years, low-resource Machine Reading Comprehension (MRC) has made significant progress, with models getting remarkable performance on various language datasets. However, none of these models have been customized for the Urdu language. This work explores the semi-automated creation of the Urdu Question Answering Dataset (UQuAD1.0) by combining machine-translated SQuAD with human-generated samples derived from Wikipedia articles and Urdu RC worksheets from Cambridge O-level books. UQuAD1.0 is a large-scale Urdu dataset intended for extractive machine reading comprehension tasks consisting of 49k question Answers pairs in question, passage, and answer format. In UQuAD1.0, 45000 pairs of QA were generated by machine translation of the original SQuAD1.0 and approximately 4000 pairs via crowdsourcing. In this study, we used two types of MRC models: rule-based baseline and advanced Transformer-based models. However, we have discovered that the latter outperforms the others; thus, we have decided to concentrate solely on Transformer-based architectures. Using XLMRoBERTa and multi-lingual BERT, we acquire an $F_1$ score of 0.66 and 0.63, respectively.

## 1. Introduction

Text comprehension and question answering remain difficult task for machines that requires large-scale resources for training. The scarcity of annotated datasets in low-resource Asian languages is one of the primary reasons the development of language-specific Question Answering models is behind, particularly in the case of the Urdu language. Some techniques for dataset creation for low-resource languages transfer English resources in order to do NLP tasks. In response to increased demand and a dearth of standard datasets in Urdu, we introduce UQuAD1.0 (Urdu Question Answering Dataset): a large-scale question-answering dataset built for Urdu MRC. We gathered 4K Urdu QA pairs via crowdsourcing and combined them with 45k Urdu translated SQuAD [1] tuples. The study includes statistics on the distribution of answers and questions, along with the types of questions. By releasing training data publicly for reading comprehension tasks, UQuAD1.0 contributes to multi-lingual language processing research. While machine translation (MT) for languages with minimal resources has proven to be a challenging task [2] [3], the level of difficulty grows furthermore when translating between two morphologically rich and morphologically poor languages [4] [5]. For that, we have developed the following research questions:

- RQ1: Can QA resources for other languages be created just by translating English resources?

- RQ2: Can time and effort be saved when manually annotating QA materials in other languages by utilizing existing resources in English?

- RQ3: Is it possible to learn Urdu RC using pre-trained multi-lingual architectures that have been trained in a variety of languages?
- RQ4: Is it possible to evaluate a model based on its language understanding capability?

In RQ1, we look at MT, or Machine-Translated Squad resource. MT should be enough to train an Urdu-QA system equivalent to SQuAD- trained even with flawless translation. However, there are two issues: (1) Translation shifts or loses the position of the answer span. (2) The quality of QA pairs varies. Without overcoming these obstacles, $F_1$ performance using our best performing model is 12.49, demonstrating the weakness of this technique. We determined from the results of RQ1 that MT performance is low, so it is appropriate to create language-specific (Urdu) resources for QA. As a result, we built the small Urdu MRC benchmark dataset using the same crowdsourcing technique that was used to build the English SQuAD, with the following contributions::

1) Including a variety of question types: To evaluate different aspects of the MRC model's language understanding capability, we provide different question types based on Bloom's taxonomy.
2) Avoiding lexical shortcuts: By imposing lexical and syntactic variety while creating query similar to benchmark SQuAD dataset.
3) Our data coverage level consists of Urdu Wikipedia articles and Tafheem (RC) worksheets of Cambridge O-level books.

Although our manually annotated dataset came from a translated resource, it needed a significant amount of time and resources to educate employees in examining and repairing translated samples. It also addressed our RQ2 that manual annotation takes the same amount of time and effort. To answer RQ3, we fine-tuned an Urdu QA system using small crowdsourced data by humans and large translated resources with a multi-lingual Transformer-based architecture, achieving an $F_1$ score of 0.66. Finally, for RQ4, we will evaluate the model's language comprehension capabilities by examining the types of questions it can answer and their associated accuracy.

This research discusses related work in section [2](), followed by dataset construction and statistics in section [3](). Later on, models are described in section [4,]() followed by Section [5](), showing the experiments and results. Finally, Section [6]() presents the conclusion and future work.

## 2. Related Work

A task where the system must answer questions about a document is called machine reading comprehension (MRC). This technique acquired significant acceptance following the publication of a large-scale Reading Comprehension (RC) dataset termed SQuAD [1] containing Over 100,000 questions on popular Wikipedia articles. The broad use of SQuAD has resulted in the formation of other related datasets. For instance, TriviaQA [6] comprises 96k questions and answers regarding trivia games, which were discovered on the Internet and documents con- training the answers. The Natural Questions corpus is a set of questions [7] that is almost three times the size of SQuAD, and the questions were extracted from Google search logs. MS MARCO [8] has one million queries extracted from Bing Search.

Unfortunately, there are very few similar MRC datasets for other languages, necessitating the development of multi-lingual MRC for low-resource languages, XSQuAD [9] dataset was built to meet this demand. It contains 40 paragraphs and 1190 question-answer pairs from SQuAD that have been translated into ten languages. Arabic and Hindi are also included in XSQuaD, but not

Urdu. In order to address the unavailability of datasets other than English RC, significant efforts have been made in recent years to develop datasets in low-resource languages for Reading Comprehension, for example, SberQuAD [10], a dataset similar to SQuAD for the Russian language, was recently created using the same technique as SQuAD. Additionally, [11] offered a Bulgarian dataset, [12] presented a Tibetan dataset, [13] generated an Arabic Reading Comprehension Dataset (ARCD) to fill the gap of MRC in other languages. Also, SQuAD-it [14], a semi-automatic translation of the SQuAD dataset into Italian, is a huge dataset for question answering processes in Italian containing 60k question/answer pairs. [15] released FQuAD: French Question Answering Dataset in two versions with 25k and 60k samples. [16] Introduced HindiRC consists of only 127 questions from 24 paragraphs, manually annotated by humans. Another variation of synthetic dataset translated from SQuAD 1.1 for Hindi reading comprehension by [17] consists of over 18k questions. The absence of native language annotated datasets other than English is one of the primary reasons language-specific Question Answering models take longer to develop.

As previously stated in the literature review, Hindi and Arabic are two low-resource languages that have evolved rapidly in MRC, with the research community giving them the attention they deserve in recent years. Although Urdu shares many characteristics with Arabic, Persian, and Hindi, such as the lack of capitalization, compound words, similar morphology, and free word order [5], Urdu still struggles to make initial studies in NLP. These languages are among the most widely spoken globally, with 170 million Urdu speakers, 490 million Hindi speakers, and 255 million Arabic speakers worldwide. These low-resource languages are highly sought after in real-world applications such as human-robot interaction, question answering, recommendations, and particular search queries. It has a knock-on impact on every business because robots that comprehend questions and react with appropriate information may boost efficiency and save time. Unfortunately, Arabic and Hindi do not have monolingual big-size MRC datasets to create state-of-the-art RC models, but they are making progress in this area by experimenting with alternate methodologies. On the other hand, the Urdu research community lags behind its close allies, as it lacks even a single dataset in the Nastaliq script.

To the best of our knowledge, the MRC contains no contributions in Urdu. To address the scarcity of Urdu language comprehension data, we present UQuAD1.0, an Urdu QA dataset for reading comprehension consisting of a total of 49k tuples (Question, paragraph, and Answer). Our research complements previous efforts by annotating small resources while utilizing large resources generated for another language. From a model architecture standpoint, most existing state-of-the-art models for reading comprehension rely on transformer-based architectures that use the self-attention mechanism to weigh the significance of each component of the passage data differently and achieve good performance.

# 3. UQuAD1.0 Dataset

UQuAD dataset consists of two main parts: a large-scale Machine Translated (MT) part and a small-scale manually annotated part using a crowdsourced approach. Both parts will be presented in detail in sections 3.1 and 3.2. General statistics about each portion are presented in Table 1.

## 3.1 Machine Translated UQuAD1.0

To address RQ1, we examine the difficulty of retaining answer spans from English to Urdu. We identify the following three examples as shown in Appendix A based on Google Translate of English SQuAD tuples into Urdu:

- Exact matching (36%): English answer spans are translated into exact Urdu terms.
- Synonym's matching (17%): The Urdu answer spans are paraphrased versions of the Urdu passage's terms.
- Unpreserved Spans (47%): Google Translation cannot retain answer spans throughout translation due to a language barrier or translation inaccuracy.

We were able to collect 53% of the UQuAD dataset using the first two approaches. However, we had to delete 47% of the data due to answer span issues. It is clear from the last example in Appendix A that a feminine pronoun is referred to as a male throughout the phrase, and in other cases, the answer was not kept between paragraph and answer owing to translation discrepancies.

## 3.2 Crowd-Source UQuAD1.0

The main challenge in training QA systems is poorly translated QA pairs. Example 3 in Appendix A shows that machine translations cannot find the correct answer span in a large portion of translated. We thus build small-scale language-specific human-generated resources for fine-tuning QA systems. The advantage of resource is near perfect precision, with the disadvantage of being labor-intensive. Similar to the SQuAD1.0 collection process, we crowdsourced over 4k question-answer pairs. The data generation procedures for this dataset are generally the same as SQuAD1.0. However, we exploited unique aspects of the Urdu language, such as extensive vocabulary usage and diversity of question types as per Bloom's taxonomy (Appendix B), to enrich and diversify this dataset. For the question-answer creation process, we recruited volunteers from different cities so that everyone had their distinct style of questioning, which added more variety to the dataset. We used a dedicated user interface (UI) guided by SQuAD guidelines to build human-generated QA pairs. We took a sample of 100 Urdu Wikipedia pages and extracted paragraphs of considerable length without graphics.

Figure 1: UQuAD UI for adding new data points

The 100 articles resulted in 1972 paragraphs covering various topics from politics, religion, education, and music, as reflected in Table 2. Human annotators utilized the UI depicted in Figure 1 to read a text, enter questions, choose types of questions specific to Bloom's taxonomy, and then highlight the spans containing answers. The practice of generating questions by copying and pasting content from Wikipedia was restricted.

## 3.3 UQuAD Question and Answer Types Analysis

To increase the difficulty of this dataset, we limit MRC models from adopting simple techniques based on fundamental word matching. Additionally, rather than focusing exclusively on keywords, our goal is to generate questions that can be addressed by examining the entire passage. Not all questions are equally challenging. Some questions are simple to answer, while others may need much thinking. Bloom [18] provides us with a taxonomy to assist in framing queries at various thinking levels. It divides cognitive abilities into six categories, ranging from low-level ability to high-level ability that requires deeper cognitive instruction. Each question is complicated in its own way, and comprehending and correctly replying to each demand a separate set of cognitive talents. In figure 5.3, we identified the types of reasoning required to solve Bloom's taxonomy-based question and presented the results of a manual assessment of 200 questions drawn from the test set. The most commonly requested question type accounts for 26.4% of all inquiries that fall under the category of Remember, which we may query using lexical variants or by rearranging synonyms. Analyze questions, which account for 19.6% of all questions, require the collection of evidence from multiple sentences. 2.9% of questions fall into the comprehend category, which interprets the message provided in the question using the various cues listed in column 3 of Appendix 2. Finally, in the external knowledge category, we checked if the response was not in the text or if the response area was erroneously picked owing to the worker's error. Similar to SQuAD dataset, we establish five forms of reasoning necessary to answer 200 questions from the test set of UQuAD, summarized in Figure 5.3. The most often requested inquiry type constitutes 27% of test data and involves rearranging the syntax or altering the phrasing of the supporting phrase. Questions from passages employing a synonym and global knowledge account for 20% and 10%. 13% of questions need proof from several sentences. On average, 10% of questions featured a deduction for a sentence's options that satisfy the question's requirements. Finally, 20% of questions were asked based on information outside the paragraph or were picked erroneously owing to a human error. For answers, we categorize UQuAD answers into six groups, shown in Table 3. It results in 18% object responses, followed by person, date, and place. Description and reasoning questions account for 17.4% and 8.3%, respectively. We conclude that UQuAD1.0 has more Date and Person classes than SQuAD1.0, but other classes are relatively equivalent.

|  |  | UQuAD (All) | UQuAD (train) | UQuAD (test) |
|---|---|---|---|---|
| Number of questions |  | 46,481 | 40,526 | 5,955 |
| Number unique paragraphs |  | 18,812 | 16,840 | 1,972 |
| avg. number of sentences per paragraph |  | 6.28 | 6.23 | 6.33 |
| Tokens | avg. paragraph length | 162.34 | 159.27 | 168.11 |
|  | Avg. question length | 12.81 | 12.75 | 12.92 |
|  | Avg. answer length | 3.75 | 3.81 | 3.48 |
| Characters | avg. paragraph length | 562.17 | 551.57 | 582.45 |
|  | Avg. question length | 43.37 | 43.17 | 43.70 |
|  | Avg. answer length | 14.86 | 15.20 | 14.27 |

Table 1: Statistics on training and development files of UQuAD1.0 dataset

| Politics | Education | Religion | Music | Misc. |
|----------|-----------|----------|-------|-------|
| 30%      | 15%       | 25%      | 18%   | 12%   |

Table 2: UQuAD1.0: Distribution of Wikipedia Articles

| Description | Reason | Person | Place | Date | Object | Time |
|-------------|--------|--------|-------|------|--------|------|
| 17.4%       | 8.3%   | 21.3%  | 11%   | 25%  | 18%    | 12%  |

Table 3: UQuAD1.0 Main Answer Types Distribution

# 4. Models

We investigate the performance of three models: a baseline approach based on sliding window [19] and two multi-lingual Transformer-based models BERT [20] and XLMRoberta [21]. The Sliding window approach was first introduced in the MCTest paper, and it solves the answer extraction problem in a rule-based manner without any training data needed. BERT is a powerful pre-trained model that recently obtained state-of-the-art performance on various NLP tasks. We use the multi-lingual pre-trained model released by Google to fine-tune the BERT model for the UQuAD1.0 task without applying additional language-specific NLP techniques. The Third model used is XLM-Roberta, a multi-lingual pre-trained transformer model from common Crawl trained on 100 languages, including Urdu. XLM-Roberta outperformed previous multi-lingual models such as mBERT and XLM on a variety of downstream tasks.

## 4.1 Sliding Window Baseline

We picked the sliding window as the baseline approach since it is also utilized in the benchmark SQuAD work and also because it demonstrates that matching term frequency or simple word matching between question and context cannot address the RC problem. For a given (paragraph, question), the sliding window approach works as follows:

1) Tokenize P-Q-A Tuple: converts each Paragraph-Question-Answer tuple to a set of tokens using a dedicated Urdu tokenizer from Stanford Stanza Library.

2) Generate Candidate Answers: generate a list of text spans of input paragraph. They are treated as candidate answers.

3) Score Candidate Answers (SW+D): Score each candidate's answer using Sliding Window and Distance features.

4) Compute Final Score: for each candidate answer. Final score = sliding window score – distance score.

5) Predict Answer: The candidate answer with the highest score is the answer predicted by the model.

Since UQuAD is not a multi-choice question dataset, we had to generate candidate answers from scratch. For that, we first generate all possible text spans from the passage with a threshold on the maximum length of candidate answers. Then we only keep answers that have the highest unigram and bigram overlap score with the related question. The sliding window and distance-based scores of each candidate answer are computed using the algorithms in Figure 3.

```
Type1: Syntactic variation (27%)
Q: What is the Rankine cycle sometimes called?
سوال: رینکین سائیکل کو بعض اوقات کیا کہا جاتا ہے؟
Context: The Rankine cycle is sometimes referred to as a practical Carnot cycle.
رینکین سائیکل کو بعض اوقات ایک عملی کارنوٹ سائیکل کہا جاتا ہے۔

Type2: Lexical variation – World Knowledge(20%)
Q Which governing bodies have governing powers?
سوال: کس انتظامیہ کے پاس حکمرانی کے اختیارات ہیں؟
Context: The European Parliament and the Council of the European Union have powers of amendment and veto during the legislative process.
یورپی پارلیمنٹ اور یورپی یونین کی ایوان کے پاس قانون سازی کے عمل میں ترمیم اور ویٹو کے اختیارات ہیں۔

Type3: Syntactic variation (10%)
Q. What Shakespeare scholar is currently on the faculty?
سوال: فیکلٹی میں اس وقت کونسے شیکسپیئر اسکالر ہیں؟
Context: Current faculty include the anthropologist Marshal Sahlins, … Shakespeare scholar David Bevington.
جملہ: موجودہ فیکلٹی میں ماہر بشریات مارشل سابلنز، ۔۔ ، شیکسپیئر اسکالر ڈیوڈ بیونگٹن شامل ہیں۔

Type4: Multiple sentences reasoning (13%)
Q. What collection does the V&A Theatre & Performance galleries hold?
سوال: وی اینڈ اے تھیٹر اینڈ پرفارمنس گیلیریز کونسا مجموعہ رکھتے ہیں؟
Context: The V&A Theatre & Performance galleries opened in March 2009. … They hold the UK's biggest national collection of material about live performance.
جملہ: وی اینڈ اے تھیٹر اینڈ پرفارمنس گیلیریز کا آغاز مارچ 2009 میں ہوا۔ وہ براہ راست پرفارمنس کے مواد کے حوالے سے برطانیہ کا سب سے بڑا قومی ذخیرہ رکھتے ہیں۔

Type5: Ambigious (10%)
Q. What is the main goal of criminal punishment?
سوال: فوجداری سزا کا بنیادی مقصد کیا ہے؟
Context: Achieving crime control via incapacitation and deterrence is a major goal of criminal punishment.
جملہ: نا اہلی اور ممانعت کے ذریعے جرائم پر قابو پانا فوجداری سزا کا بنیادی مقصد ہے۔
```

Figure 2: Examples of UQuAD reasoning abilities Similar to SQuAD, we manually categorized 200 tuples into one or more above groups. The Crowdsourced solution is underlined, and words relating to the reasoning type are highlighted.

## 4.2 Transformer models: XLMRoberta and mBERT

In this work, the performance of transformer-based models for machine-reading comprehension is examined. Models built on top of the Transformer architecture [22] account for the vast majority of state-of-the-art performance in a variety of natural language processing tasks. In the absence of pre-trained Urdu monolingual models, we leverage the transfer learning concept inherent in these models to fine-tune the current pre-trained models for question answering on the UQuAD dataset. We examined various such models and found out that two of them outperform all others in our experiments: multi-lingual BERT (mBERT) and XLM-RoBERTa. They already incorporated knowledge of 104 and 100 languages, respectively, including Urdu.

## 4.3 Train/Validation/Test split

We created two sets of annotated QA pairs, one for training (80%) and one for testing (20%), with no overlap of passages or articles. Statistics on both portions are presented in Table 1. The training part was furthermore split into 80% for actual training and 20% for validation. We use 5-fold cross-validation for the better significance of the performance results.

### 4.3.1 Data Preprocessing

While mBERT and XLMRoBERTa are distinct models, their overall fine-tuning process using the Transformers library is similar, demonstrating the API's potential. There- fore, we describe in the following paragraphs the common foundation for both models. We begin by preprocessing the whole dataset to eliminate any noise or inconsistency introduced during data gathering. We eliminate instances that exhibit one or more of the following:

- The paragraph does not have an answer span.
- Index of incorrect responses (i.e., paragraph text at answer index is different from answer text).
- The index of the response is -1.

We then determine the model based on each response's start/end indices in the resulting data. We know the answer's placement in terms of its character index inside the paragraph, but we require its location in terms of the model's internal tokenization system. To do this, we proceed as follows for each (Paragraph, Question, Answer) tuple:

1) Tokenize the response to ascertain the number of tokens it contains.

2) In the associated paragraph, replace the response with a list of the model's mask tokens (i.e., [MASK] for mBERT and <mask> for XLM- RoBERTa), according to the number of tokens in the answer. Following that, tokenize the result.

3) Determine the start/end indices of the response in the tokenized paragraph by locating the mask tokens in the encoded result.

---

**Baseline Algorithms**
**Require:** Passage $P$, set of passage words $PW$, $i^{th}$ word in passage $P_i$, set of words in question $Q$, set of words in hypothesized answers $A_{1..4}$, and set of stop words $U$,
**Define:** $C(w) := \sum_i \mathbb{I}(P_i = w)$;
**Define:** $IC(w) := \log\left(1 + \frac{1}{C(w)}\right)$.

---

**Algorithm 1 Sliding Window**
for $i = 1$ to 4 do
$\quad S = A_i \cup Q$
$\quad sw_i = \max_{j=1..|P|} \sum_{w=1..|S|} \begin{cases} IC(P_{j+w}) & \text{if } P_{j+w} \in S \\ 0 & \text{otherwise} \end{cases}$
end for
return $sw_{1..4}$

---

**Algorithm 2 Distance Based**
for $i = 1$ to 4 do
$\quad S_Q = (Q \cap PW) \setminus U$
$\quad S_{A_i} = ((A_i \cap PW) \setminus Q) \setminus U$
$\quad$ if $|S_Q| = 0$ or $|S_{A_i}| = 0$
$\quad\quad d_i = 1$
$\quad$ else
$\quad\quad d_i = \frac{1}{|P|-1} \min_{q \in S_Q, a \in S_{A_i}} d_P(q, a)$,
$\quad\quad$ where $d_P(q, a)$ is the minimum number of words between an occurrence of $q$ and an occurrence of $a$ in $P$, plus one.
$\quad$ end if
end for
return $d_{1..4}$

Figure 3: Sliding Window and Distance-based Algorithms

### 4.3.2 Fine-Tuning Process

The maximum input sequence length (max_len) at which samples will be truncated is a small but essential step. It is essential for successful fine-tuning because using the longest input sequence as the threshold would slow down the training process and potentially cause memory overflow, resulting in Out Of Memory (OOM) failures. Both the maximum length and batch size must fit inside the memory constraints of our GPU. As a result, there is a tradeoff between data loss, memory usage, and training speed. To do this, we calculate data loss as a percentage of the maximum length value and tolerate up to 2% data loss. Then we choose a threshold that considers all three variables (i.e., data loss, training speed, memory usage). Finally, we load the pre-trained model, set its hyperparameters (learning rate, optimizer, batch size), and begin fine-tuning the model using the previously encoded UQuAD training data. For each model, we trained two sub-models using the same architecture: one to predict the response start index and the other to predict the answer end index.

### 4.3.3 Hyperparameters

Table 4 presents the values of the hyperparameters used during fine-tuning. A maximum input sequence length of 384 and batch size of 16 is the highest combination the available GPU (i.e., Tesla T4) could handle without memory issues, with data loss smaller than 2% and acceptable speed (1h30min per epoch). We use Adam optimizer [23] (more specifically, the AdamW variant and a dynamic learning rate that decreases linearly over training steps, with a number of warmup steps=0. The validation set is a randomly sampled 20% of the training dataset used to tune hyperparameters. The number of epochs was tuned by plotting the training and validation loss for different numbers of epochs (ref Figure 4). During fine-tuning, the training loss keeps decreasing while the validation initially decreases then starts increasing after several epochs. It reflects the start of over-fitting. We choose the previous epoch as the best epoch. The remaining hyperparameters values were chosen according to the standard values recommended by the model's authors in their papers. We furthermore investigate changes in a set of hyperparameters values without performance improvement.

| Hyperparameter | Value |
| --- | --- |
| Learning Rate | 2e-5 |
| Nbr. Epochs | 2 |
| Maximum Input Length | 384 |
| Batch Size | 16 |
| Optimizer | Adam W |
| Adam's Epsilon | 1e-8 |

Table 4: Hyperparameters values used during Fine-tuning

# 5. Experiments and Results

## 5.1 Model Evaluation

In many aspects, the MRC work is comparable to a human reading comprehension task in terms of complexity. As a result, MRC model evaluation can take the same form: the model responds to paragraph queries and is assessed by comparing the model's replies to the right ones. This gives an answer to RQ3 about the model's capability of learning RC. We may compare the model's output to the right answer and assign it a score of 1 if they are identical and 0 if they are not. This metric is called Exact Match. This, on the other hand, will consider answers that are partially accurate as wrong responses. Even if the model's output is incredibly near to the correct answer, the exact match score will still be zero if the correct answer is "KotAdu city" and the model output is "KotAdu " the exact match score will still be zero.

Consequently, the $F_1$ metric, which is the harmonic mean of the accuracy and recall, is often used for extracting responses. Precision and recall of a model are measured by the proportion of words in the model's output that appear in the correct response, and recall is measured by the proportion of words in the correct response that appear in the model's output. Precision is measured by the proportion of words in the correct response in the model's output. The $F_1$ measure can only give a partial score when the model's output is only partially correct due to this limitation. Figure 5 provides an example of calculating the EM and $F_1$ scores and explains how to do so.

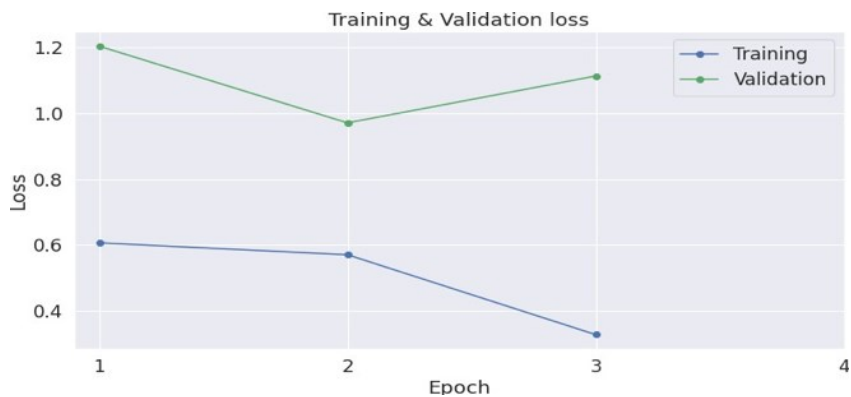

Figure 4: Tuning number of epochs for mBERT. After epoch two, the validation loss starts increasing while the training loss keeps decreasing (over-fitting). The optimal number of epochs is 2

## 5.2  Results

We assess the performance of the three models on the UQuAD test set 23% of the questions in the test set to feature more than one possible answer, which provides greater versatility when assessing the model, as the same question may have many answers, which may occur in a variety of locations across the paragraph. The evaluation is carried out using two widely used measures for Machine Reading Comprehension: Accuracy/Exact Match (EM) metric and $F_1$ score. The $F_1$ score indicates the average overlap between the predicted response and the true answer, whereas the EM represents the proportion of predicted answers precisely matching accurate answers. While the Sliding Window base model achieved decent performance on English SQuAD (i.e., $F_1$ score of 0.2), its application to Urdu did not have good results. The model achieved a very low accuracy of 4% and an $F_1$ score of 0.03. Given its lexical particularities, this rule-based algorithm focusing primarily on matching words and distance between them is underperforming when applied to the Urdu language. When testing the Transformer-based models XLMRoBERTa and mBERT, we select the answer start index with the highest probability in the dedicated sub-model (i.e., the sub-model that predicts the answer start index).

Similarly, we take the predicted response end- index from the dedicated sub-model. Since each sub-model was trained independently of the other, we evaluate its performance individually. The final model performance is the average of both sub-models performances. Both XLMRoBERTa and mBERT achieved excellent performance with $F_1$ scores of 0.66 and 0.63, respectively.

|  | Sliding window | mBERT | XLM-RoBERTa |
|---|---|---|---|
| $F_1$ | 3% | 63% | 66% |
| Exact Match | 4% | 66% | 36% |

Table 5: Models Performance Results using EM and $F_1$ score

However, their Exact Match scores are significantly lower, and this is due to the model predicting answers with a high level of words intersection but slightly away from an exact match. For example, a 5words predicted answer of which four words match precisely the answer is considered a wrong answer and accounts for 0 in the EM, while the $F_1$ score that uses word-based evaluation will be high. We investigate this aspect by calculating the percentage of predicted answers in the actual answers and vice-versa. We found for XLMRoBERTa (resp. mBERT) that 47% (resp. 50%) of the predicted answers are in the associated accurate answers, and 66% (i.e., 17%) of the actual answers are in the associated predicted answers. It also compares the whole answers with the same words order; dropping the words order results in higher percentages, thus the high $F_1$ scores achieved by the models. Table 5 summarizes the performance evaluation findings—the high performance of the transformer-based model's semantic aspects of Urdu at a broader level. Also, the self-attention mechanism helps memorize relevant parts of the paragraphs that are key to extracting the final answer. Both XLMRoBERTa and mBERT were pre-trained on text corpora of millions of documents from different languages. This transfer learning process reduces the fine-tuning time and required data volume and benefits the model from understanding general aspects of different languages that might apply in a wide range of use cases, especially for relatively similar languages such as Arabic and Hindi share a large set of common characteristics.

Figure 5: Evaluation example using Exact Match and $F_1$ score

## 5.3    MRC Performance stratified by a question and answer types

This part will assess the language comprehension power and limitations of our most accurate model, XLM-RoBERTa, using three criteria: Urdu Question Difficulty, named-entities, and question type (who, what, when, where, and which). The proportion of accurately predicted responses (i.e., an exact match between predicted and real answers) for each question (resp. answer) type is shown in Figure 6(a) (resp. (b)). We only display question/answer types for which there are more than ten occurrences in the test set in both charts. We may see in (a) that performance on "What" questions is poorer than on other WH questions. It corresponds with our understanding because the "what" question is not as explicit as "Where" for the place, "When" for the date, and "Who" for an individual. The "Which" Question likewise received a poor score. In (b), the model gets 97% percent of the country answers correct. We anticipate that this is due to the transfer learning process, as the countries would have been met repeatedly during pre-training in multiple languages.

Moreover, because they were few as compared to dates and locations, the model quickly grasped them. When compared to others, performance on organization and person sorts of answers is relatively poor. A thorough error analysis of these questions reveals that the model can comprehend the context of the question and provide a sentence-containing answer despite being "ambiguous questions" in nature. While the model's forecast for "Exact Match" questiontypes is essentially true even although the term "services"is not retrieved, even though the question words are paraphrases of context terms, the model was able to provide an accurate solution to the "Synonym Matching Questions." As seen in Appendix C, it was able to connect the word "legislative process" to the "governing body" through the use of word knowledge. For anaphora-based questions, the model cannot distinguish the antecedent "Chaudhry Aitzaz Ahsan" from the subsequent"He. " An interesting approach would be to test if the model can handle multi-sentence reasoning challenges and cataphora resolution issues in the absence of relevant training samples. The overall results are compelling and correspond to the model's predicted behavior. However, this is far from a thorough exploration of the model's comprehension, and a more in-depth examination of the model's explainability might provide fascinating results.

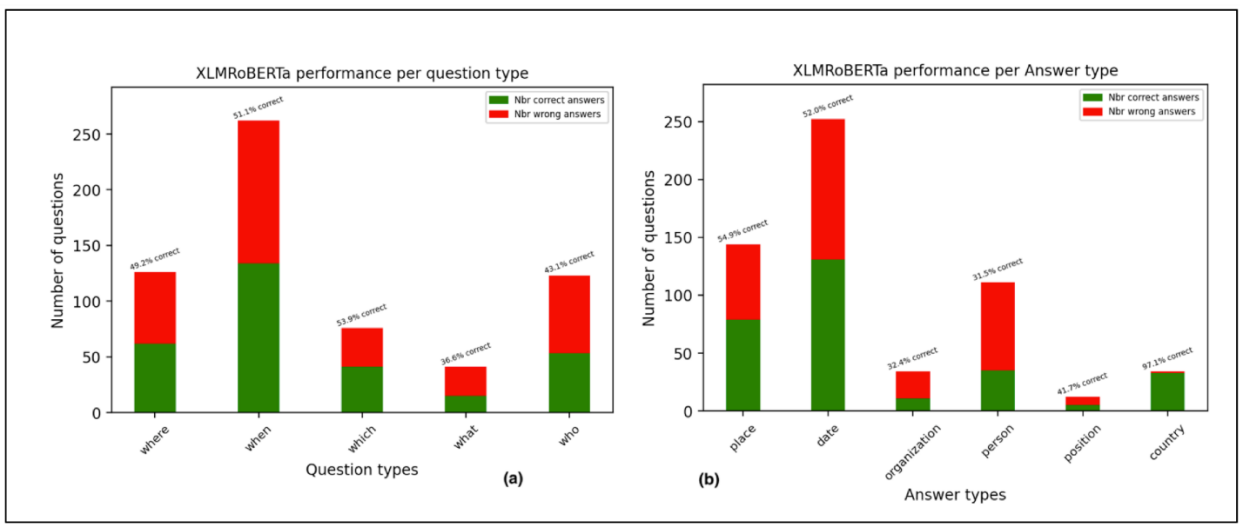

Figure 6: XLMRoBERTa Performance per Question/Answer Type

# 6. Conclusion and Future Work

This study proposes UQuAD1.0, a new large-scale, open-domain Urdu machine reading comprehension (MRC) dataset that is intended to address real-world MRC problems of Urdu-low resource language. UQuAD provides the following advantages over datasets from other low-resource languages akin to Urdu: (1) **Data sources**: Automatic translation of original SQuAD1.0, human-generated questions and answers from Wikipedia pages, and Cambridge O-level books (2) **Diverse Question Groups**: It provides question annotations on test set based on three distinct groupings: wh questions, named entities, and Bloom's taxonomy. We fine-tuned a baseline model on the test set to get an $F_1$ score of 0.66 and an exact match ratio of 0.36 . After thoroughly evaluating the model's result, we concluded that the model was unable to handle descriptive and anaphora-related questions. Further study should be conducted to discover the model's limitations and develop specialized additions that are particular to the Urdu language. In order to bridge the performance gap between training state-of-the-art models in Urdu MRC, we are making baseline systems and data public.

## Acknowledgments

The authors would like to acknowledge their crowd-working and translation team for their work on the UQuAD1.0 dataset. The authors would also like to thank Dr. Hassan Sajjad, Hamza Lebbar, Furqan Shaikh, and the anonymous reviewers for their feedback and comments.

# Appendix A

**Examples of Translation Categories From English SQuAD to Urdu**

| English | Urdu Translation |
|---|---|
| **Type1: Exact Match** | |
| **P:** Ghulam Ishaq Khan Bangash (February 22, 1915 - October 27, 2006) was a former President of Pakistan. He served in government positions long before he entered politics. He was born into a Pashtun family in Ismail Khel, a village in Bannu District. He belonged to the Bangash tribe of Pashtuns. After his primary education, he graduated from Peshawar with a degree in Chemistry and Botany. Joined the Indian Civil Service in 1940 | غلام اسحاق خان بنگش (22 فروری، 1915ء تا 27 اکتوبر، 2006ء) پاکستان کے سابق صدر تھے۔ انہوں نے سیاست میں آنے سے بہت پہلے سرکاری عہدوں پر خدمات سر انجام دیں۔ ضلع بنوں کے ایک گاؤں اسماعیل خیل میں ایک پشتون گھرانے میں پیدا ہوئے۔ ان کا تعلق پشتونوں کے بنگش قبیلے سے تھا ابتدائی تعلیم کے بعد انہوں نے پشاور سے کیمسٹری اور باٹنی کے مضامین کے ساتھ گریجویشن کی۔ انیس سو چالیس میں انڈین سول سروس میں شمولیت اختیار کی |
| **Q:** What did Ghulam Ishaq Khan do before entering politics? | سیاست میں آنے سے پہلے غلام اسحاق خان کیا کام کرتے تھے ؟ |
| **A:** Services in government positions | سرکاری عہدوں پر خدمات |
| **Type2: Synonymy Matching** | |
| **P:** Beyoncé is an American singer, songwriter, record producer and actress. Born and raised in Houston, Texas, she done in various singing and dancing competitions as a child, and rose to fame in the late 1990s as lead singer of R&B girl-group Destiny's Child. Managed by her father, Mathew Knowles, the group became one of the world's best-selling girl groups of all time. Their hiatus saw the release of Beyoncé's debut album, Dangerously in Love (2003), which proven her as a solo artist worldwide, received five Grammy Awards and featured the Billboard Hot 100 number-one singles "Crazy in Love" and "Baby Boy". | ایک امریکی گلوکار، نغمہ نگار ، اور ریکارڈ پروڈیوسر اداکارہ ہیں۔ ہیوسٹن ، ٹیکساس میں پیدا ہوا اور بڑا ہوا، اس نے بچپن میں مختلف گانے اور رقص کے مقابلوں میں پرفارم کیا، اور 1990 کی دہائی کے آخر میں آر اینڈ بی گرل گروپ ڈیسٹنیز چائلڈ کی مرکزی گلوکارہ کی حیثیت سے شہرت حاصل کی۔ اس کے والد ، میتھیو نولس کے زیر انتظام ، یہ گروپ اب تک کی دنیا کی سب سے زیادہ فروخت ہونے والی لڑکیوں کے گروہوں میں سے ایک بن گیا۔ ان کے وقفے نے بیونے کی پہلی البم ، ڈینجرسلی ان ان لیو (2003) کی ریلیز دیکھی ، جس نے اسے دنیا بھر میں سولو آرٹسٹ کے طور پر قائم کیا، پانچ گریمی ایوارڈ حاصل کیے اور بل بورڈ ہاٹ 100 نمبر ون سنگلز "کریزی ان لو" اور "بیبی بوائے" کو نمایاں کیا۔ |
| **Q:** In what city and state was Beyonce born and raised? | بیونس کس شہر اور ریاست میں پیدا ہوا اور بڑا ہوا؟ |
| **A:** born and raised in Houston, Texas | یوسٹن ، ٹیکساس میں پیدا ہوئے اور پرورش پائی۔ |
| **Type3: Unpreserved Span** | |
| **P:** Following the disbandment of Destiny's Child in June 2005, she released her second solo album, B'Day (2006), which contained hits "Déjà Vu", "Irreplaceable", and "Beautiful Liar". Beyoncé also ventured into acting, with a Golden Globe-nominated performance in Dreamgirls (2006), and starring roles in The Pink Panther (2006) and Obsessed (2009). Her marriage to rapper Jay Z and portrayal of Etta James in Cadillac Records (2008) influenced her third album, I Am... Sasha | جون 2005 میں تقدیر کے بچے کو ختم کرنے کے بعد، اس نے اپنا دوسرا واحد البم B'Day (2006) جاری کیا، جس میں "Déjà Vu" اور ، "ناقابل جگہ"، "خوبصورت جھوٹا" تھا۔ بیونے نے ڈریم گرلز (2006) میں گولڈن گلوب نامزد کارکردگی ، اور دی پنک پینتھر (2006) اور آبروسید (2009) میں مرکزی کردار ادا کرتے ہوئے اداکاری میں بھی جتن کیا۔ ریپر جے زیڈ کے ساتھ اس کی شادی اور کیڈیلک ریکارڈز (2008) میں اٹا جیمس کی تصویر کشی نے ان کی تیسری البم ، آئی ایم ... ساشا فیئرس (2008) کو متاثر کیا، جس نے ان کی تبدیل شدہ انا ساشا فیئرس کی پیدائش |
| Fierce (2008), which saw the birth of her alter-ego Sasha Fierce and earned a record-setting six Grammy Awards in 2010, including Song of the Year for "Single Ladies (Put a Ring on It)". Beyoncé took a hiatus from music in 2010 and took over management of her career; her fourth album 4 (2011) was subsequently mellower in tone, exploring 1970s funk, 1980s pop, and 1990s soul. Her critically acclaimed fifth studio album, Beyoncé (2013), was distinguished from previous releases by its experimental production and exploration of darker themes. | دیکھی اور ریکارڈ قائم کیا۔ 2010 میں چھ گریمی ایوارڈ، جس میں "سنگل خواتین (اس پر رنگ لگائیں)" کے گانے کا سال شامل ہے۔ بیونسے نے سن 2010 میں میوزک سے وقف کر لیا تھا اور اپنے کیرئیر کا انتظام سنبھال لیا تھا۔ اس کا چوتھا البم 4 (2011) بعد میں لہجے میں مختاط تھا، جس نے 1970 کی دہائی کی فنک، 1980 کی پاپ اور 1990 کی دہائی کی روح کو تلاش کیا۔ اس کے تنقیدی طور پر سراہا جانے والا پانچواں اسٹوڈیو البم بیونس (2013)، کو اس کی تجرباتی تیاری اور گہرے موضوعات کی تلاش کے ذریعے پچھلی ریلیز سے ممتاز کیا گیا تھا۔ |
| **Q:** For what movie did Beyonce receive her first Golden Globe nomination? | بیونس نے کس فلم کے لئے پہلی گولڈن گلوب کی نامزدگی حاصل کی؟: |
| **A:** Dreamgirls | خوابوں والی لڑکیاں |

# Appendix B
**Types of questions based on Bloom's taxonomy in UQuAD Dataset**

| Category | Definition | Question Cues | Question Example |
|---|---|---|---|
| Remember | Recalling factual information from long term memory like dates, event etc. | کون، کب، کہاں، کتنے شناخت، نام، فہرست بنائیں، حوالہ، کس | پاکستان کا بانی کون ہے؟<br>امریکہ کا یوم آزادی کب ہے؟<br>پیرس کہاں واقع ہے؟<br>پاکستان کے کتنے صوبے ہیں؟<br>نیوزی لینڈ کے قومی پرندے کی شناخت کریں؟<br>آپ کا نام کیا ہے؟<br>محمد علی جناح کے 14 نکات کی فہرست بنائیں؟<br>قرآن مجید میں کون سی اخلاقی اقدار کا حوالہ دیا گیا ہے؟<br>علی کس یونیورسٹی سے پڑھتا ہے؟ |
| Understand | Interpret the instructional messages like oral, written, and graphical. | فرق، بیان، خلاصہ، تبادلہ خیال، پیش گوئی، برعکس، تشریح، موازنہ، غور، بحث، بحال کریں، اجاگر | ہارڈ ویئر اور سافٹ ویئر میں کیا فرق ہے؟<br>پاکستان کا نظریہ بیان کریں؟<br>راؤنڈ ٹیبل کانفرنس کے نقطہ کا خلاصہ کریں؟<br>موسمیاتی تبدیلی کے اثرات پر تبادلہ خیال کریں؟<br>آج کے موسم کی پیش گوئی کیا ہے؟<br>جمہوریت اور آمریت کے درمیان موازنہ اور برعکس کریں؟<br>اس نظم کی تشریح کریں؟<br>تعلیم میں اخلاقی اقدار کی اہمیت پر غور/ بحث کریں؟<br>تعلیم کے چار روایتی فلسفہ مقاصد کو نمایاں کریں؟<br>چارٹر سکولوں کے فوائد اور نقصانات کو اجاگر کریں؟ |
| Apply | Implementing or executing procedures in real life scenario. | مظاہرہ کریں، حساب، حل، مثال دیں، کیسے، جائزہ، اطلاق، معائنہ، استعمال | اس مشین کو کیسے استعمال کریں؟<br>فائر ڈرل کے طریقہ کار کا مظاہرہ کریں؟<br>اس مسئلے کو کیسے حل کیا جائے؟<br>ان دو دیہات کے درمیان فاصلے کا حساب لگائیں؟<br>ہماری روز مرہ زندگی میں ٹیکنالوجی کے استعمال کی مثالیں دیں؟<br>مشین کی فعالیت کا جائزہ لیں؟<br>مشین اے اور مشین بی کے درمیان فرق کا معائنہ کریں؟<br>آپ نے اس کہانی سے کیا سیکھا؟ |

| Analyse | Dividing document into its constituent parts to identify how parts are related to each other and overall material and its purpose. | تجزیہ کریں، وضاحت کریں، درجہ بندی، تقسیم، تعلق، کیوں، کیا، | کمپیوٹر کو سائز کی بنیاد پر درجہ بندی کریں؟ تعلیمی فلسفہ کے مختلف مکتبہ فکر کا تجزیہ کریں؟ موجودہ تعلیمی نظام میں تعلیم کے اسلامی فلسفہ وضاحت کریں؟ پودوں کو ان کے مسکن کی بنیاد پر تقسیم کریں؟ موسمیاتی تبدیلی اور گلوبل وارمنگ کے درمیان کیا تعلق ہے؟ جسم میں دل کا کام کیا ہے؟ |
|---|---|---|---|
| Evaluate | Judging, critiquing and checking according to certain criteria. | جانچ پڑتال، جانچ، رسائی، درجہ، گریڈ، منتخب کریں، پیمانہ، قائل، موثر، تحقیقات، فیصلہ، تلاش، اندازہ، جواز پیش کریں، | یہ مشین کتنی موثر ہے؟ آپ اس نظام کو کیا درجہ دیں گے؟ نوجوان اخبار کے مطالعے سے لاتعلق کیوں ہیں؟ حیاتیاتی تنوع کے ناپید ہونے کے ذمہ دار عوامل کی تحقیقات کریں؟ نظام کی کارکردگی کی جانچ پڑتال کریں؟ اس مقابلے کے نتائج کا فیصلہ کریں؟ درج ذیل ریاضی کے مسئلے میں غلطیاں تلاش کریں؟ فلم "دی پیٹریاٹ" کی درستگی کا اندازہ کریں؟ کیا فنون سکول کے نصاب کا ایک اہم حصہ ہیں؟ اپنے جواب کا جواز پیش کریں؟ |
| Create | Combining element to generate novel coherent whole or producing an original product. | ڈیزائن، اختراع، مفروضہ، ، کرافٹ، ایجاد، بنایا، تحریر، کمپوز، تجویز، منصوبہ، بنائیں، تیار | گاڑی کا بلیو پرنٹ ڈیزائن کریں؟ صنعتی انقلاب کے موجدوں کے بارے میں ایک نیا بورڈ گیم ایجاد کریں؟ موسیقی کا ایک نیا ٹکڑا تحریر کریں جس میں سی میجر کی کلید میں راگ شامل ہوں؟ کوئی نیا میوزک کمپوز کریں؟ لنچ روم میں طلباء کو اپنے بعد صفائی کروانے کا کوئی متبادل طریقہ تجویز کریں؟ نوعمر سگریٹ نوشی روکنے میں مدد کے لیے ایک مہم کا منصوبہ بنائیں؟ ایک بل بنائیں جو آپ اسمبلی میں پاس ہوتے دیکھنا چاہیں گے؟ سائنس فیئر پروجیکٹ کے لیے ایک آئیڈیا تیار کریں جو پودوں کی زندگی پر آلودگی کے اثرات پر مرکوز ہو؟ |

# Appendix C
**Qualitative results of the top-performing model on different question groups**

| English | UrduTranslation |
|---|---|
| **Type1:Exact Match** | |
| **P**:Ghulam Ishaq Khan Bangash (February 22, 1915 - October 27, 2006) was a former President of Pakistan. He served in government positions long before he entered politics. He was born into a Pashtun family in Ismail Khel, a village in Bannu District. He belonged to the Bangash tribe of Pashtuns. After his primary education, he graduated from Peshawar with a degree in Chemistry and Botany. Joined the Indian Civil Service in 1940 | غلام اسحاق خان بنگش (22 فروری، 1915ء تا 27 اکتوبر، 2006ء) پاکستان کے سابق صدر تھے۔ انہوں نے سیاست میں آنے سے بہت پہلے سرکاری عہدوں پر خدمات سر انجام دیں۔ ضلع بنوں کے ایک گاؤں اسماعیل خیل میں ایک پشتون گھرانے میں پیدا ہوئے۔ ان کا تعلق پشتونوں کے بنگش قبیلے سے تھا۔ ابتدائی تعلیم کے بعد انہوں نے پشاور سے کیمسٹری اور باٹنی کے مضامین کے ساتھ گریجویشن کی۔ انیس سو چالیس میں انڈین سول سروس میں شمولیت اختیار کی |
| **Q**: What did Ghulam Ishaq Khan do before entering politics? | سیاست میں آنے سے پہلے غلام اسحاق خان کیا کام کرتے تھے ؟ |
| **Gold Answer** : Services in government positions | سرکاری عہدوں پر خدمات |
| **Predicted Answer** | سرکاری عہدوں پر |
| **Type2:  Synonymy Matching** | |
| **P**: Beyoncé is an American singer, songwriter, record producer and actress. Born and raised in Houston, Texas, she done in various singing and dancing competitions as a child, and rose to fame in the late 1990s as lead singer of R&B girl-group Destiny's Child. Managed by her father, Mathew Knowles, the group became one of the world's best-selling girl groups of all time. Their hiatus saw the release of Beyoncé's debut album, Dangerously in Love (2003), which proven her as a solo artist worldwide, received five Grammy Awards and featured the Billboard Hot 100 number-one singles "Crazy in Love" and "Baby Boy". | ایک امریکی گلوکار، نغمہ نگار، اور ریکارڈ پروڈیوسر اداکارہ ہیں۔ ہیوسٹن، ٹیکساس میں پیدا ہونے اور بڑا ہوا، اس نے بچپن میں مختلف گانے اور رقص کے مقابلوں میں پرفارم کیا، اور 1990 کی دہائی کے آخر میں آر اینڈ بی گرل گروپ ڈیسٹنیز چائلڈ کی مرکزی گلوکارہ کی حیثیت سے شہرت حاصل کی۔ اس کے والد، میتھیو نولس کے زیر انتظام، یہ گروپ اب تک کی دنیا کی سب سے زیادہ فروخت ہونے والی لڑکیوں کے گروہوں میں سے ایک بن گیا۔ ان کے وقفے نے بیوٹے کی پہلی البم، ڈینجرسلی ان لو (2003) کی ریلیز دیکھی، جس نے اسے دنیا بھر میں سولو آرٹسٹ کے طور پر قائم کیا، پانچ گریمی ایوارڈ حاصل کیے اور بل بورڈ ہاٹ 100 نمبر ون سنگلز "کریزی ان لو" اور "بیبی بوائے" کو نمایاں کیا۔ |
| **Q**: In what city and state was Beyonce born and raised? | بیونے کس شہر اور ریاست میں پیدا ہوا اور بڑا ہوا؟: |
| **Gold Anaswer**:born and raised in Houston, Texas | ہیوسٹن ، ٹیکساس میں پیدا ہوئے اور پرورش پائی۔ |
| **Predicted Answer**: | اسماعیل |
| **Syntactic Variation** | |
| **P**:The Rankine cycle is sometimes **referred** to as a **practical Carnot cycle**.. | رینکین سائیکل کو بعض اوقات ایک عملی کارنوٹ سائیکل کہا جاتا ہے۔ |
| **Q**: What is the Rankine cycle sometimes **called**? | رینکین سائیکل کو بعض اوقات کیا کہا جاتا ہے ؟ |
| **Gold Answer**: **practical Carnot cycle**. | عملی کارنوٹ سائیکل |
| **Predicted Answer**: | عملی کارنوٹ |
| **Lexical variation – World Knowledge** | |
| **P: The European Parliament and the Council of the European Union** have powers of amendment and veto during the legislative process. | یوپی پارلیمنٹ اور یوپی یونین کی ایوان کے پاس قانون سازی کے عمل میں ترمیم اور ویٹو کے اختیارات ہیں۔ |
| **Q**:Which **governing bodies** have governing powers | کس انتظامیہ کے پاس حکمرانی کے اختیارات ہیں؟ |
| **Gold Answer:** The European Parliament and the Council of the European Union | یوپی پارلیمنٹ اور یوپی یونین |
| **Predicted Answer:** | یورپی پارلیمنٹ اور یورپی یونین |
| **Ambiguous** | |
| **P:** Achieving crime control via <u>incapacitation</u> and <u>deterrence</u> is a major goal of criminal punishment. | نا اہلی اور ممانعت کے ذریعے جرائم پر قابو پانا فوجداری سزا کا بنیادی مقصد ہے |
| **Q:** What is the main goal of criminal punishment? | فوجداری سزا کا بنیادی مقصد کیا ہے؟ : |
| **Gold Answer** | جرائم پر قابو پانا |
| **Predicted Answer** | نا اہلی اور ممانعت کے ذریعے جرائم پر قابو پا |
| **Named Entities (Wh Question for answering named entities—here wh question keyword is underlined)** | |
| The Islamic Republic of Pakistan is an independent Islamic country located in the defensive part of South Asia, Northwest Central Asia, and West Asia. With a population of 21 crores, it is the fifth most populous country in the world. With 881,913 square kilometers (340,509 square miles), it is the 33rd largest country in the world. It is bounded on the south by 1,046 km (650 miles) of coastline that joins the Arabian Sea. Pakistan is bordered by India to the east, China to the northeast, and Afghanistan and Iran to the west. | اسلامی جمہوریہ پاکستان جنوبی ایشیا کے شمال مغرب وسطی ایشیا اور مغربی ایشیا کے لیے دفاعی طور پر اہم حصے میں واقع ایک خود مختار اسلامی ملک ہے۔ 21 کروڑ کی آبادی کے ساتھ یہ دنیا کا پانچواں بڑی آبادی والا ملک ہے۔ 881,913 مربع کلومیٹر (340,509 مربع میل) کے ساتھ یہ دنیا کا تینتیسواں بڑے رقبے والا ملک ہے۔ اس کے جنوب میں 1046 کلومیٹر (650 میل) کی ساحلی پٹی ہے جو بحیرہ عرب سے ملتی ہے۔ پاکستان کے مشرق میں بھارت، "شمال مشرق میں چین اور مغرب میں افغانستان اور ایران واقع ہیں۔ - |
| **Q:** What is the total population of Pakistan? | پاکستان کی آبادی <u>کتنی</u> ہے؟ |
| **Gold Answer:** 21 crores | 21 کروڑ |
| **Predicted Answer** | 21 کروڑ |
| **Named Entities (who, which, how many, where)** | |
| **P:** William Shakespeare was an English writer and poet who is considered one of the greatest English writers and playwrights in the world. Shakespeare is generally called the National Poet of England. His writings include about 38 games, 154 songs, two long poems, and many short poems. His games have been translated into all the major languages of the world and are presented more than any other author's games. Shakespeare was Born in Stratford. Not much is known about his early life. | ولیم شیکسپیئر ایک انگریز مصنف اور شاعر تھا جسے انگریزی زبان میں دنیا کے عظیم ترین مصنفین اور ڈراما نگاروں میں شمار کیا جاتا ہے۔ شیکسپیئر کو عموماً انگلستان کا قومی شاعر کہا جاتا ہے۔ اس کی تصانیف میں تقریباً 38 کھیل، 154 گیت دو لمبی نظمیں اور بہت سی چھوٹی نظمیں شامل ہیں۔ اس کے کھیل کی سبھی دنیا کی بڑی زبانوں میں ترجمہ ہوچکے ہیں اور کسی بھی دوسرے مصنف کے کھیلوں کی نسبت زیادہ پیش کیے جاتے ہیں۔ شیکسپیئر سٹراٹ فورڈ میں پیدا ہوئے۔ ان کی ابتدائی زندگی کے بارے میں زیادہ معلومات دستیاب نہیں ہیں۔ - |
| **Q:** Who was William Shakespeare? | ولیم شیکسپیئر کون تھا؟ |
| **Gold Answer:** writer and poet | مصنف اور شاعر |
| **Predicted Answer** | مصنف |
| **Q:** William Shakespeare is the national poet of which country? | ولیم شیکسپیئر کو کس ملک کا قومی شاعر کہا جاتا ہے؟ |
| **Gold Answer:** England | انگلستان |
| **Predicted Answer:** England | انگلستان |
| **Q:** How many songs did William Shakespeare write? | ولیم شیکسپیئر نے کتنے گیت لکھے ہیں؟ |
| **Gold Answer:** 154 songs | 154 گیت |
| **Predicted Answer:** | 154 گیت |
| **Q:** Where was Shakespeare born? | شیکسپیئر کہاں پیدا ہوئے؟ |
| **Gold Answer:** Stratford | سٹراٹ فورڈ |

| Predicted Answer | سٹرائٹ فورڈ |
|---|---|
| **Anaphora Resolution** | |
| P: Chaudhry Aitzaz Ahsan Pakistan's is a renowned lawyer and politician. Barrister in Law Chaudhry Aitzaz Ahsan was born on September 27, 1945 in Murree, Rawalpindi District. He studied at Aitchison College, Lahore for his primary education, and then at Government College, Lahore. He later moved to Cambridge University in the United Kingdom to study law. On his return from Cambridge, Aitzaz Ahsan took the CSS exam. He sharply criticized the government of then military dictator General Ayub Khan | چوہدری اعتزاز احسن پاکستان کے معروف قانون دان اور سیاستدان ہیں۔ بیرسٹر ان لاء چوہدری اعتزاز احسن 27 ستمبر 1945 کو مری، ضلع راولپنڈی میں پیدا ہوئے۔ انہوں نے اپنی ابتدائی تعلیم ایچی سن کالج لاہور سے حاصل کی اور پھر گورنمنٹ کالج لاہور سے تعلیم حاصل کی۔ بعد میں وہ قانون کی تعلیم حاصل کرنے کے لیے برطانیہ کی کیمبرج یونیورسٹی چلے گئے۔ کیمبرج سے واپسی پر اعتزاز احسن نے سی ایس ایس کا امتحان دیا۔ انہوں نے اس وقت کے فوجی آمر جنرل ایوب خان کی حکومت پر کڑی تنقید کی۔ |
| **Q:** Whose government did Chaudhry Aitzaz Ahsan sharply criticize at that time? | چودھری اعتزاز احسن نے اس وقت کس کی حکومت کے خلاف شدید تنقید کی؟ |
| **Gold Answer:** General Ayub Khan | جنرل ایوب خان |
| **Predicted Answer** | NoAnswer |
| Q: Where did Chaudhry Aitzaz Ahsan stay studying for elementary education? | چوہدری اعتزاز احسن ابتدائی تعلیم کے لیے کہاں زیر تعلیم رہے؟ |
| **Gold Answer:** Aitchison College | ایچی سن کالج |
| **Predicted Answer** | NoAnswer |